\documentclass[10pt,conference]{IEEEtran}
\IEEEoverridecommandlockouts
\usepackage{cite}
\usepackage{amsmath,amssymb,amsfonts}
\usepackage{algorithmic}
\usepackage{graphicx}
\usepackage{textcomp}
\usepackage{xcolor}
\def\BibTeX{{\rm B\kern-.05em{\sc i\kern-.025em b}\kern-.08em
    T\kern-.1667em\lower.7ex\hbox{E}\kern-.125emX}}

\usepackage{url}
\usepackage{graphicx}
\usepackage{booktabs}
\usepackage{caption}
\usepackage{subfigure}
\usepackage{makecell}  

\makeatletter
\renewcommand\normalsize{%
\@setfontsize\normalsize\@xpt\@xiipt
\abovedisplayskip 3\p@ \@plus3\p@ \@minus5\p@
\abovedisplayshortskip \z@ \@plus3\p@
\belowdisplayshortskip 6\p@ \@plus3\p@ \@minus3\p@
\belowdisplayskip \abovedisplayskip
\let\@listi\@listI}
\makeatother

\begin{document}

\title{How Robust is Federated Learning to Communication Error? A Comparison Study Between Uplink and Downlink Channels}
\author{\IEEEauthorblockN{Linping Qu$^*$, Shenghui Song$^*$, Chi-Ying Tsui$^*$, and Yuyi Mao$^\dag$}
\IEEEauthorblockA{$^*$Dept. of ECE, The Hong Kong University of Science and Technology, Hong Kong\\
$^{\dag}$Dept. of EEE, The Hong Kong Polytechnic University, Hong Kong\\
Email: lqu@connect.ust.hk, eeshsong@ust.hk, eetsui@ust.hk, yuyi-eie.mao@polyu.edu.hk}
}

\maketitle

\begin{abstract}
Because of its privacy-preserving capability, federated learning (FL) has attracted significant attention from both academia and industry. However, when being implemented over wireless networks, it is not clear how much communication error can be tolerated by FL. This paper investigates the robustness of FL to the uplink and downlink communication error. Our theoretical analysis reveals that the robustness depends on two critical parameters, namely the number of clients and the numerical range of model parameters. It is also shown that the uplink communication in FL can tolerate a higher bit error rate (BER) than downlink communication, and this difference is quantified by a proposed formula. The findings and theoretical analyses are further validated by extensive experiments.
\end{abstract}


\section{Introduction}
\label{Introduction}
Federated learning (FL) is a distributed machine learning scheme that can train a global model based on local data at different clients, without violating their data-privacy requirements\cite{mcmahan2017communication}. However, because frequent exchanges of deep neural network (DNN) models between the clients and server over noisy wireless channels are needed, the communication bottleneck becomes one of the biggest challenges for FL\cite{ang2020robust}. 

Existing works address the communication challenge in FL through techniques such as model quantization \cite{reisizadeh2020fedpaq,tonellotto2021neural,amiri2020federated}, sparsification \cite{han2020adaptive,ozfatura2021time,li2020ggs}, and low-rank approximation\cite{konevcny2016federated,zhou2020low}, with an assumption of error-free communication channels. However, communication error is inevitable and will significantly affect both learning and communication performance of FL. While some previous research has tried to address this issue by focusing on the impact of communication error, it has been limited to analog communication\cite{yang2020federated,xia2021fast,sery2021over}.
Some recent studies like \cite{amiri2020federated}, \cite{amiri2021convergence}, and \cite{jiang2020cluster} have investigated FL in digital communication, considering digital modulation or quantizing model parameters into bit streams. However, the impact of communication error on FL over digital communication networks has not been fully understood. 

To overcome the communication error and achieve a low bit error rate (BER), high transmitting power and channel decoding effort are usually required. However, due to the inherent error tolerance of DNN\cite{hasan2019tolerance}, FL may absorb a certain BER without learning accuracy degradation, which indicates an opportunity to reduce the energy consumption in wireless FL. To exploit this opportunity, we need to understand the tolerance of FL to BER, which unfortunately remains unknown. To fill this research gap, in this paper, we investigate the robustness of FL to BER that exists in both the uplink and downlink communication.

The contributions of this paper are as follows:\\
\noindent1) We investigate the robustness of FL to BER in the uplink and downlink. After theoretical analyses, we identify the critical parameters that contribute to the error-tolerating capacity.\\
\noindent2) Based on the theoretical results, we find that the FL uplink is more tolerant to BER than the downlink because of model aggregation and a narrower range of gradients, and we come up with a formula to quantify this difference in error tolerance.\\
\noindent3) The critical parameters, the superiority of the uplink in BER tolerance, and the formula to quantify the difference are all validated by experiments. The validated theoretical results will be useful in the design of wireless FL.

The remainder of this paper is organized as follows. In Section II, we describe the system model that captures the BER in the uplink and downlink of FL. In Section III, we present theoretical analyses to reveal what contributes to the BER tolerance in the downlink and uplink. In Section IV, we provide experimental results to validate our theory. Finally, we conclude our paper in Section V.

\section{System Model}
\label{System Model}
In this section, we first briefly introduce FL, and then model how communication error affects it.

\subsection{Federated Learning}
\label{Federated Learning}
We consider a horizontal FL system that comprises one server and $n$ clients (e.g., mobile devices), which collaborate to train a DNN model through multiple communication rounds. At each round, the global model is broadcasted to clients, updated on the local data, and then transmitted to the server to produce a new global model by aggregation.

Specially, federated learning solves the following optimization problem \cite{li2020federated}:
\begin{align}
{\underset{\mathbf{w}}{\rm min} \ f(\mathbf{w}) = \underset{\mathbf{w}}{\rm min}\sum\nolimits_{i=1}^np_if_i(\mathbf{w})},
\end{align}
where $\mathbf{w}\in {\mathbb{R}^d}$ represents the DNN model, with $d$ denoting its dimension; $p_i$ is the ratio of data size at the $i^{th}$ device to that of all devices; and $f_i(\mathbf{w})$ denotes the local loss function of the $i^{th}$ device.

\subsection{Federated Learning with Communication Error}
\label{Federated Learning with BER}
\begin{figure}[t]
    \begin{center}
    \includegraphics[width=0.7\linewidth]{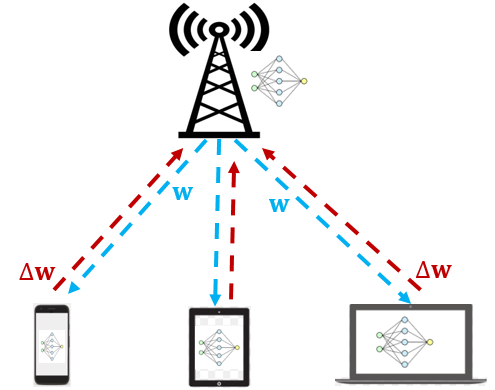}
    \end{center}
    \caption{FL over noisy wireless channels.}
    \label{fig:system model}
\end{figure}

As depicted in Fig.~\ref{fig:system model}, both the uplink and downlink communication channels between the clients and the server are noisy, which introduces communication error to model parameters.
Denoting the global model at the server in the $m^{th}$ communication round as $\mathbf{w}_m$, the received model at mobile devices for the first step of local training is a noisy version of the global model, and can be given as
\begin{align}
&\mathbf{w}^i_{m,0}=\mathbf{w}'_m,
\label{Bw}
\end{align}
where $\mathbf{w}'_m$ denotes the global model with BER impact.

Then the $i^{th}$ client conducts $\tau$ steps of local stochastic gradient descent (SGD) with a learning rate $\eta$. The local updating rule is given by
\begin{align}
\mathbf{w}^i_{m,t+1}=\mathbf{w}^i_{m,t}-\eta{\nabla} f_i(\mathbf{w}^i_{m,t}),
\label{receive}
\end{align}
where $t=0, ..., \tau-1$, and ${\nabla} f_i(\mathbf{w}^i_{m,t})$ represents the stochastic gradient computed from the local training datasets.
Upon completing $\tau$ steps of local training, the $i^{th}$ client obtains a new model $\mathbf{w}^i_{m,\tau}$. The local model update can be obtained as
\begin{align}
{\Delta \mathbf{w}^i_m=\mathbf{w}^i_{m,\tau}-\mathbf{w}^i_{m,0}}.
\end{align}
Then, the local model update is transmitted to the server through the noisy uplink communication channel. Again, what the server receives is a noisy version of the local model update, given by $(\Delta \mathbf{w}^i_m)'$. Upon receiving the noisy local updates from the mobile devices, the server performs aggregation to obtain an updated global model using the following equation:
\begin{align}
\mathbf{w}_{m+1}=\mathbf{w}_{m}+\sum\nolimits_{i=1}^np_i(\Delta \mathbf{w}^i_m)'.
\end{align}
The above process iterates until the algorithm converges.

\section{BER Tolerance: Uplink vs. downlink}
\label{BER Tolerance: Uplink and downlink}
In this section, we first intuitively analyze the difference in BER tolerance between the downlink and uplink. Then we provide theoretical analyses which can explain the difference and also quantify the BER tolerance with parameters.

\subsection{Intuitive Analysis}
\label{Intuitive Analysis}
\begin{figure}[t]
    \begin{center}
    \includegraphics[width=0.7\linewidth]{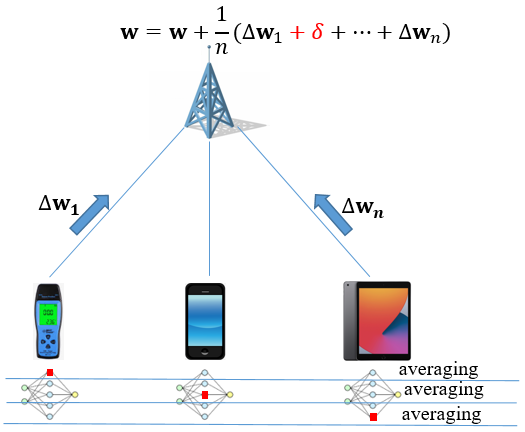}
    \end{center}
    \caption{The uplink error can be averaged in the server.}
    \label{fig:uplink averaging}
\end{figure}

Intuitively, we expect that the uplink can tolerate a higher BER than the downlink, for two reasons. First, for a given BER in the downlink, the local training steps on a noisy model may propagate the errors. However, for the uplink, the effect of the given BER may be alleviated by the aggregation operation. As shown in Fig.\ref{fig:uplink averaging}, due to the randomness, the error bits in different clients are likely to appear in different DNN positions. When aggregated on the server side, the error value appearing in the given position may be averaged by the number of clients. Therefore, the error is reduced in the uplink.

Second, considering a typical FL system that broadcasts model weights in the downlink and transmits model updates in the uplink, the range of the model weights is usually larger than the range of model updates. The BER appearing in the parameters with a larger range will translate into a bigger error in magnitude, so the given BER may have a severer impact on the downlink.

\subsection{Theoretical Analysis}
\label{Theoretical Analysis}
To theoretically analyze the BER tolerance, we list the following basic assumptions \cite{suresh2017distributed, reisizadeh2020fedpaq}:\\
\textbf{Assumption 1.} \textit{The loss function $f_i$ is L-smooth with respect to $\mathbf{w}$, i.e., for any $\mathbf{w}$ and $\hat{\mathbf{w}} \in \mathbb{R}^d$, we can have $||\nabla f_i(\mathbf{w})-\nabla f_i(\hat{\mathbf{w}})||\leq L||\mathbf{w}-\hat{\mathbf{w}}||$.}\\
\textbf{Assumption 2.} \textit{Stochastic gradients $\tilde{\nabla} f_i(\mathbf{w})$ are unbiased and variance bounded, i.e., $\mathbb{E}_{\xi}[\tilde{\nabla} f_i(\mathbf{w})]=\nabla f_i(\mathbf{w})$, and $\mathbb{E}_{\xi}[||\tilde{\nabla} f_i(\mathbf{w})-\nabla f_i(\mathbf{w})||^2] \leq \sigma^2$, where $\xi$ denotes the mini-batch dataset, and $\sigma^2$ denotes the variance.}

Using the above assumptions, we can provide the convergence bound for wireless FL with a given BER in the uplink or downlink. To investigate the extreme BER tolerance for one given link, we assume no error for the other link.\\
\noindent\textbf{Theorem 1.} \textit{Considering the wireless FL system introduced in Section II with non-convex loss functions and a total of $K$ communication rounds, the convergence bound of FL with a downlink BER is
\begin{align}
&\frac{1}{K\tau}\sum_{m=0}^{K-1}\sum_{t=0}^{\tau-1}\mathbb{E}\Vert \nabla f(\bar{\mathbf{w}}_{m,t})\Vert^2  \nonumber\\
&\leq \frac{2Ld}{3K\tau\eta}\sum_{m=0}^{K-1}BER_m\cdot range^2(\mathbf{w}_m) \nonumber\\
&+\frac{2(f(\mathbf{w}_0)-f^*)}{K\tau\eta} +\frac{L^2(n+1)(\tau-1)\eta^2\sigma^2}{n}+\frac{L\eta\sigma^2}{n},
\label{Theorem_downlink}
\end{align}
and the convergence bound of FL with an uplink BER is
\begin{align}
&\frac{1}{K\tau}\sum_{m=0}^{K-1}\sum_{t=0}^{\tau -1}\mathbb{E}\Vert \nabla f(\bar{\mathbf{w}}_{m,t})\Vert^2 \nonumber\\
& \leq \frac{Ld}{3n^2K\tau\eta}\sum_{m=0}^{K-1}\sum_{i \in [n]}BER_m^i\cdot range^2(\Delta \mathbf{w}_m^i) \nonumber \\
&+\frac{2(f(\mathbf{w}_0)-f^*)}{K\tau\eta}+\frac{L^2(n+1)(\tau-1)\eta^2\sigma^2}{n}+\frac{L\eta\sigma^2}{n}, 
\label{Theorem_uplink}
\end{align}\\
where $range()$ is equal to the maximum value in the model subtracting the minimum value; $f^*$ denotes the final converged training loss; $BER_m$ and $BER_m^i$ are the BERs in the downlink and uplink communication, respectively.}\\
\textit{Proof.} See Appendix.

On the right-hand side (RHS) of Eq.(\ref{Theorem_downlink}) or Eq.(\ref{Theorem_uplink}), the first term captures the impact of the downlink BER or uplink BER. The second term captures the distance between the initial and the final training loss. The third term shows the deviation of each local model from the overall average model, and the fourth term comes from the variance of the local gradient estimator. It is worth noting that the only difference between downlink-noisy and uplink-noisy FL is the first term.

\begin{figure}[t]
\begin{minipage}[b]{.48\linewidth}
  \centering
  \centerline{\includegraphics[width=4.0cm]{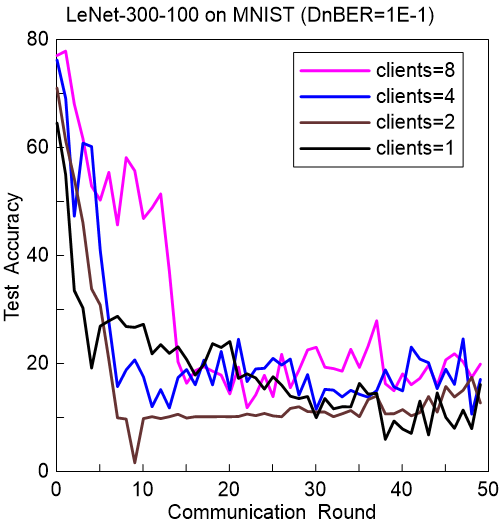}}
  \begin{center}
  (a) FL with downlink errors: Increasing the number of clients does not help training.
  \end{center}
\end{minipage}
\hfill
\begin{minipage}[b]{0.48\linewidth}
  \centering
  \centerline{\includegraphics[width=4.0cm]{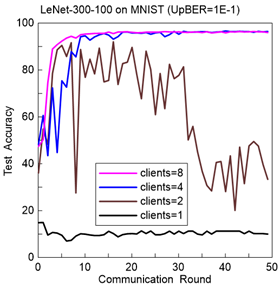}}
  \begin{center}
  (b) FL with uplink errors: Increasing the number of clients improves training.
  \end{center}
\end{minipage}
\caption{How the learning performance with downlink or uplink bit errors is affected by the number of clients.}
\label{fig: number of users}
\end{figure}

\noindent\textbf{Remark 1 (The effect of the number of clients).} The impact of BER on FL convergence can be examined by looking at the first term on the RHS of both Eq.(\ref{Theorem_downlink}) and Eq.(\ref{Theorem_uplink}). We observe that for the uplink, the impact of the BER is related to $n$, i.e., $\frac{1}{n^2}\sum_{i\in[n]}$. According to the law of large numbers, the overall effect is close to the impact of uplink BER being divided by $n$. This observation is consistent with our intuitive analysis that the impact of the uplink BER can be reduced by averaging. In contrast, for the downlink, the impact of the downlink BER has no relation to $n$.

\noindent\textbf{Remark 2 (Quantifying the difference in BER tolerance).} To quantify the difference in BER tolerance between the uplink and downlink, we can derive the required BERs for a given learning performance. Making the RHS of Eq.(\ref{Theorem_downlink}) equal to the RHS of Eq.(\ref{Theorem_uplink}), we get
\begin{align}
BER_m = \frac{BER_m^i}{2n}\left(\frac{range(\Delta \mathbf{w}_m^i)}{range(\mathbf{w}_m)}\right)^2. 
\label{relation_updates}
\end{align}
This indicates that the difference in BER tolerance between the downlink and uplink can be quantified by a function of the number of clients $n$ and the range of communicated parameters, which is also consistent with our intuitive analysis that the range of transmitted parameters matters. If considering transmitting the model weights rather than the model updates in the uplink, the Eq.(\ref{relation_updates}) will become
\begin{align}
BER_m = \frac{BER_m^i}{2n}\left(\frac{range(\mathbf{w}_m^i)}{range(\mathbf{w}_m)}\right)^2.
\label{relation_weights}
\end{align}

\begin{figure}[t]
\begin{minipage}[b]{.48\linewidth}
  \centering
  \centerline{\includegraphics[width=4.0cm]{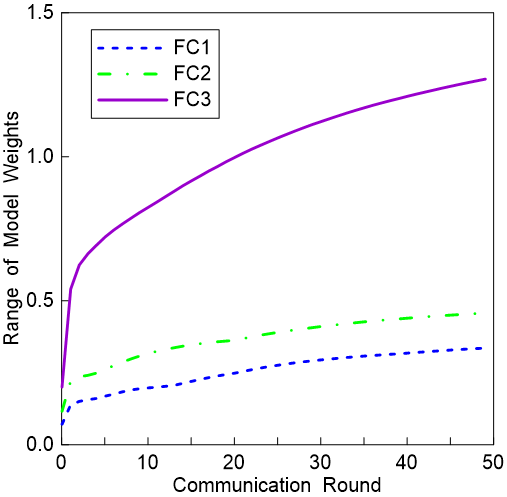}}
  \begin{center}
  (a) Range of downlink $\mathbf{w}$
  \end{center}
\end{minipage}
\hfill
\begin{minipage}[b]{0.48\linewidth}
  \centering
  \centerline{\includegraphics[width=4.0cm]{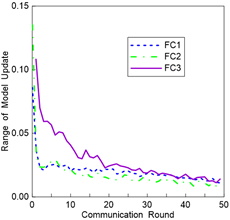}}
  \begin{center}
  (b) Range of uplink $\Delta\mathbf{w}$
  \end{center}
\end{minipage}
\caption{The range of model parameters for MNIST experiment.}
\label{fig: range mnist updates}
\end{figure}

\begin{figure}[t]
\begin{minipage}[b]{.48\linewidth}
  \centering
  \centerline{\includegraphics[width=4.0cm]{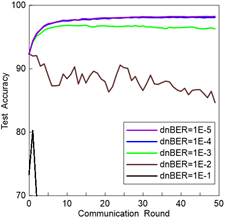}}
  \begin{center}
  (a) FL with downlink BER
  \end{center}
\end{minipage}
\hfill
\begin{minipage}[b]{0.48\linewidth}
  \centering
  \centerline{\includegraphics[width=4.0cm]{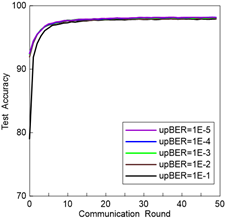}}
  \begin{center}
  (b) FL with uplink BER
  \end{center}
\end{minipage}
\caption{MNIST experiment: downlink tolerates BER of $10^{-4}$ while uplink can tolerate $10^{-1}$.}
\label{fig: dnw_upg_mnist}
\end{figure}

\begin{figure}[t]
\begin{minipage}[b]{.48\linewidth}
  \centering
  \centerline{\includegraphics[width=4.0cm]{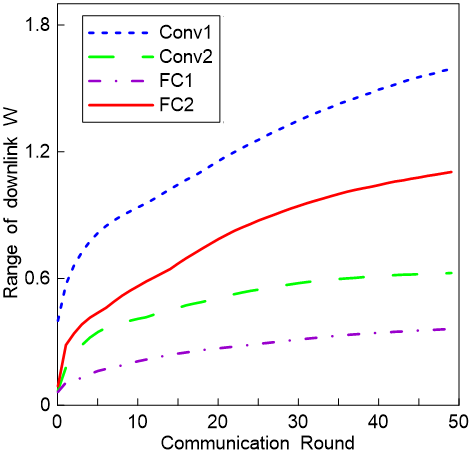}}
  \begin{center}
  (a) Range of downlink $\mathbf{w}$
  \end{center}
\end{minipage}
\hfill
\begin{minipage}[b]{0.48\linewidth}
  \centering
  \centerline{\includegraphics[width=4.0cm]{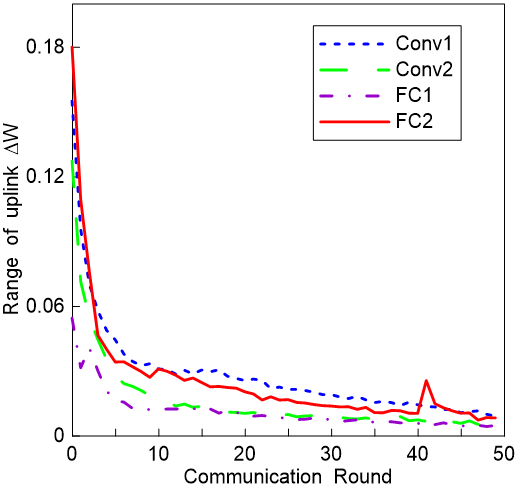}}
  \begin{center}
  (b) Range of uplink $\Delta\mathbf{w}$
  \end{center}
\end{minipage}
\caption{The range of model parameters in Fashion-MNIST.}
\label{fig: range fashion updates}
\end{figure}

\begin{figure}[t]
\begin{minipage}[b]{.48\linewidth}
  \centering
  \centerline{\includegraphics[width=4.0cm]{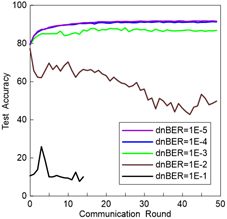}}
  \begin{center}
  (a) FL with downlink BER
  \end{center}
\end{minipage}
\hfill
\begin{minipage}[b]{0.48\linewidth}
  \centering
  \centerline{\includegraphics[width=4.0cm]{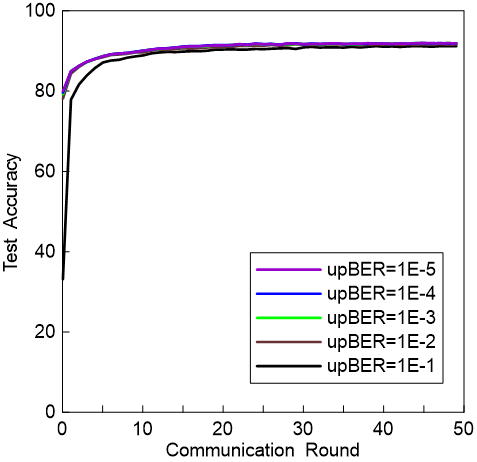}}
  \begin{center}
  (b) FL with uplink BER
  \end{center}
\end{minipage}
\caption{Fashion-MNIST experiment: downlink tolerates BER of $10^{-4}$ while uplink can tolerate $10^{-1}$.}
\label{fig: dnw_upg_fashion}
\end{figure}

From the theory, the uplink can tolerate a bigger BER than the downlink, and its superiority can be quantified by the formula in Eq.(\ref{relation_updates}) or Eq.(\ref{relation_weights}), depending on whether model updates or model weights are transmitted in the uplink. These analyses can be utilized to design wireless FL.

\section{Experiments and Discussions}
\label{Experiments and Discussions}

\subsection{Experiment Setup}
\label{Experiment Setup}
We verified our theoretical analysis from two experiments: 
1) MNIST\cite{lecun1998mnist} dataset classified by the LeNet-300-100 model\cite{han2015deep}, which is a fully connected neural network consisting of two hidden layers and one softmax output layer and
2) Fashion-MNIST\cite{xiao2017fashion} dataset classified by a vanilla CNN \cite{mcmahan2017communication}, which comprises two 5×5 convolution layers, one fully connected layer, and one softmax output layer.

The training datasets in the MNIST and Fashion-MNIST were divided among all mobile devices in an identical and independently distributed (i.i.d) manner, while the test datasets were utilized to perform validation on the server side. The local training was configured with a learning rate of 0.01, batch size of 64, and momentum of 0.5, using the SGD optimizer. The number of local training steps was set to 5.

The model parameters transmitted during the uplink and downlink communication were binarized into bit streams and some bits were randomly selected to flip to simulate the effect of BER. All experiments were conducted on a computer equipped with a $6^{th}$ Gen Intel(R) Xeon(R) W-2245 @3.90GHz CPU and NVIDIA GeForce RTX 3090 GPU.

\subsection{ Results and Analysis}
\label{ Results and Analysis}

\noindent\textbf{Validation of Remark 1 in Section III}: 
To validate Remark 1 on the effect of the number of clients, we conducted experiments on the MNIST dataset. The BER in the downlink or uplink is set the same, i.e., $10^{-1}$, and we check whether the learning performance is improved by increasing the number of clients. For FL with a downlink BER, as shown in Fig.\ref{fig: number of users}(a), increasing the number of clients from 1 to 8 does not improve the learning performance. This validates the first observation in  Remark 1 that the impact of downlink BER has no relation to the number of clients.
For FL with an uplink BER, as shown in Fig.\ref{fig: number of users}(b), increasing the number of clients from 1 to 8 gradually improves the learning performance. This validates the second observation in Remark 1 that the impact of uplink BER can be alleviated by the number of clients.

\noindent\textbf{Validation of Remark 2 in Section III}: To validate Remark 2, the number of clients was fixed to 5. We conducted experiments on the MNIST and Fashion-MNIST datasets, and considered transmitting updates and weights in the uplink.

1) To validate Eq.(\ref{relation_updates}), we transmit model updates in the uplink. For the MNIST experiment, Fig.\ref{fig: range mnist updates} shows that the range of the downlink model weights is almost 10 times the range of the uplink model updates. Then we have $range(\mathbf{w}_m) \approx 10\cdot range(\Delta \mathbf{w}_m^i)$, and Eq.(\ref{relation_updates}) becomes $BER_m \approx \frac{BER_m^i}{200n}=\frac{BER_m^i}{1000}$. As shown in Fig.\ref{fig: dnw_upg_mnist}, the BER tolerance on downlink is $10^{-4}$, while on the uplink it is $10^{-1}$, which validates Eq.(\ref{relation_updates}). The above findings also hold for the Fashion-MNIST experiment. Fig.\ref{fig: range fashion updates} shows the range of the downlink model weights and uplink model updates, and Fig.\ref{fig: dnw_upg_fashion} shows the BER tolerance, which further validates Eq.(\ref{relation_updates}).

\begin{figure}[t]
\begin{minipage}[b]{.48\linewidth}
  \centering
  \centerline{\includegraphics[width=4.0cm]{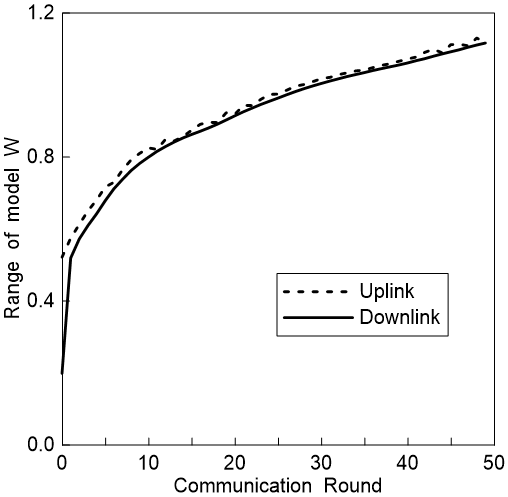}}
  \begin{center}
  (a) MNIST
  \end{center}
\end{minipage}
\hfill
\begin{minipage}[b]{0.48\linewidth}
  \centering
  \centerline{\includegraphics[width=4.0cm]{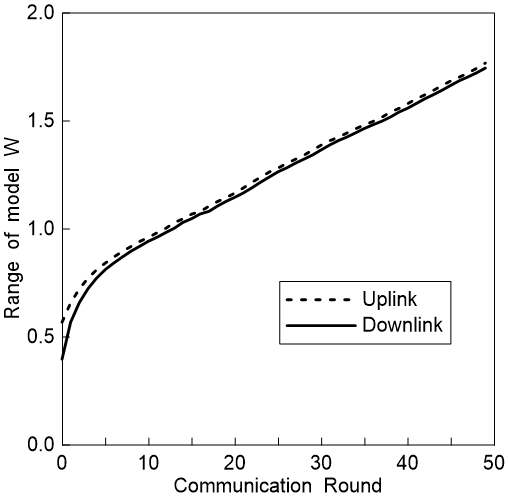}}
  \begin{center}
  (b) Fashion-MNIST
  \end{center}
\end{minipage}
\caption{For different communication rounds, the range of model weights in uplink is always similar to that in downlink.}
\label{fig: range_dnw_upw}
\end{figure}

\begin{figure}[t]
\begin{minipage}[b]{.48\linewidth}
  \centering
  \centerline{\includegraphics[width=4.0cm]{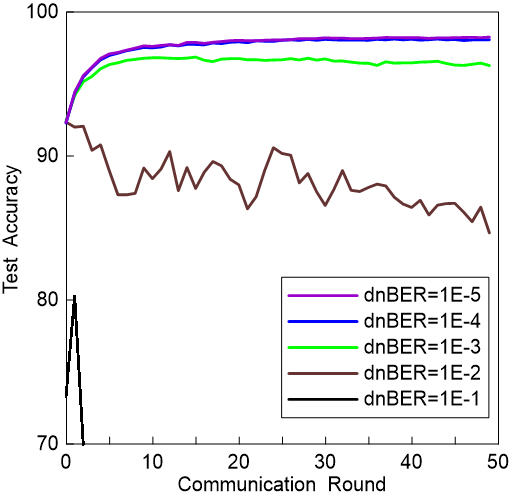}}
  \begin{center}
  (a) FL with downlink BER
  \end{center}
\end{minipage}
\hfill
\begin{minipage}[b]{0.48\linewidth}
  \centering
  \centerline{\includegraphics[width=4.0cm]{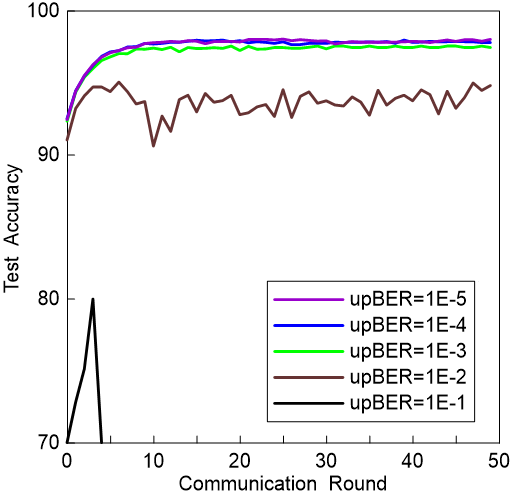}}
  \begin{center}
  (b) FL with uplink BER
  \end{center}
\end{minipage}
\caption{MNIST experiment: downlink tolerates BER of $10^{-4}$ and uplink tolerates $10^{-3}$.}
\label{fig: dnw_upw_mnist}
\end{figure}

\begin{figure}[t]
\begin{minipage}[b]{.48\linewidth}
  \centering
  \centerline{\includegraphics[width=4.0cm]{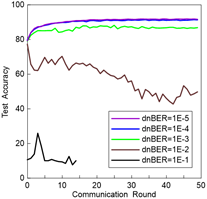}}
  \begin{center}
  (a) FL with downlink BER
  \end{center}
\end{minipage}
\hfill
\begin{minipage}[b]{0.48\linewidth}
  \centering
  \centerline{\includegraphics[width=4.0cm]{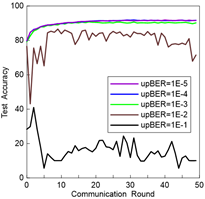}}
  \begin{center}
  (b) FL with uplink BER
  \end{center}
\end{minipage}
\caption{Fashion-MNIST experiment: downlink tolerates BER of $10^{-4}$ and uplink tolerates $10^{-3}$.}
\label{fig: dnw_upw_fashion}
\end{figure}

2) To validate Eq.(\ref{relation_weights}), we transmit model weights in the uplink. For the MNIST experiment, Fig.\ref{fig: range_dnw_upw}(a) illustrates that the range of model weights in the downlink is nearly identical to that in the uplink. Thus, Eq.(\ref{relation_weights}) simplifies to $BER_m \approx \frac{BER_m^i}{2n}=\frac{BER_m^i}{10}$. As depicted in Fig.\ref{fig: dnw_upw_mnist}, the BER tolerance in the downlink is $10^{-4}$, while in the uplink, it is $10^{-3}$. The BER tolerance in experimental results satisfies Eq.(\ref{relation_weights}). The Fashion-MNIST experiment yields similar results. Fig.\ref{fig: range_dnw_upw}(b) shows the range of model weights, and Fig.\ref{fig: dnw_upw_fashion} demonstrates the BER tolerance. These results also confirm the validity of Eq.(\ref{relation_weights}).

\section{Conclusion}
\label{Conclusion}
In this paper, we investigated the tolerance of FL to communication errors in digital communication systems. By analyzing the impact of BER on FL convergence, it was shown that the uplink is more robust to BER than the downlink due to the aggregation among a number of participating clients and the smaller range of model updates as compared to the downlink model weights. Based on theoretical analysis, we quantified the tolerance difference between the uplink and downlink communication. Experimental results on two popular datasets validated the theoretical analysis. Given that a lower requirement on the BER will significantly reduce the energy cost of a communication system, the analysis in this paper can be leveraged to design more energy-efficient FL systems.

\section{Acknowledgment}
\label{Acknowledgment}
This work was supported in part by the Research Grants Council under the Areas of Excellence scheme grant AoE/E-601/22-R, and the Hong Kong University of Science and Technology (HKUST) Startup Fund under Grant R9249.

\section*{Appendix}
\subsection{Proof of Theorem 1}
We first prove the convergence bound for FL with uplink bit errors. For communication round $m=0,1,..., K-1$ and local iteration $t=0,1,...,\tau-1$, we denote
\begin{align}
&\mathbf{w}_{m+1}=\mathbf{w}_m+\frac{1}{n}\sum_{i\in{[n]}}(\mathbf{w}_{m,\tau}^{i}-\mathbf{w}_m)',\nonumber\\
&\bar{\mathbf{w}}_{m,t}=\frac{1}{n}\sum_{i\in{[n]}}\mathbf{w}_{m,t}^{i},
\label{add_notation}
\end{align}
where $\mathbf{w}_{m+1}$ denotes the updated global model in the server, and $\bar{\mathbf{w}}_{m,t}$ denotes the averaged model of $n$ devices.\\
\textbf{Lemma 1.} If Assumptions 1 and 2 hold, then we have
\begin{align}
\mathbb{E}f(\mathbf{w}_{m+1})\leq &\mathbb{E}f(\bar{\mathbf{w}}_{m,\tau})+\frac{L}{2}\mathbb{E}\Vert \mathbf{w}_{m+1}-\bar{\mathbf{w}}_{m,\tau}\Vert^2, 
\label{lemma1}
\end{align}
which shows the property of $L$-smooth functions. The proof is provided in Appendix-B.
In the following, we derive the upper bound for the two terms in the RHS of (\ref{lemma1}).\\
\textbf{Lemma 2.} Let Assumptions 1 and 2 hold. For perfect downlink transmission, we have
\begin{align}
\mathbb{E}&f(\bar{\mathbf{w}}_{m,\tau}) \leq \mathbb{E}f(\mathbf{w}_m)-\frac{\eta}{2}\sum_{t=0}^{\tau -1}\mathbb{E}\Vert \nabla f(\bar{\mathbf{w}}_{m,t})\Vert^2 \nonumber\\
&-\eta(\frac{1}{2n}-\frac{L\eta}{2n}-\frac{L^2\eta^2\tau(\tau-1)}{n})\sum_{t=0}^{\tau -1}\sum_{i \in [n]}\mathbb{E}\Vert \nabla f(\mathbf{w}_{m,t}^{i})\Vert^2 \nonumber\\
&+\frac{L\tau\eta^2\sigma^2}{2n}+\frac{L^2\eta^3\sigma^2(n+1)\tau(\tau-1)}{2n},
\label{lemma2}
\end{align}
which captures the receiving and local training process. The proof is similar to Section 8.3 in \cite{reisizadeh2020fedpaq}. \\
\textbf{Lemma 3.} Under Assumption 1, we can obtain 
\begin{align}
\mathbb{E}\Vert \mathbf{w}_{m+1}-\bar{\mathbf{w}}_{m,\tau}\Vert^2 \leq \frac{d}{3n^2}\sum_{i \in [n]}BER_m^i\cdot range^2(\Delta \mathbf{w}_m^i),
\end{align}
which captures the model error by the uplink bit errors. The proof can be found in Appendix-B.

By introducing Lemmas 2 and 3 into Lemma 1, we can get below a recursive inequality for the global model:
\begin{align}
\mathbb{E}&f(\mathbf{w}_{m+1}) \leq \mathbb{E}f(\mathbf{w}_m)-\frac{\eta}{2}\sum_{t=0}^{\tau -1}\mathbb{E}\Vert \nabla f(\bar{\mathbf{w}}_{m,t})\Vert^2 \nonumber\\
&-\frac{\eta}{2n} \left\{1-L\eta-2\tau(\tau-1)L^2\eta^2\right\} \sum_{t=0}^{\tau -1}\sum_{i \in [n]}\mathbb{E}\Vert \nabla f(\mathbf{w}_{m,t}^{i})\Vert^2 \nonumber\\
&+\frac{L}{2n^2}\sum_{i \in [n]}\frac{d\cdot BER_m^i \cdot range^2(\Delta \mathbf{w}_m^i)}{3}\nonumber \\
&+\frac{L^2(n+1)\tau(\tau-1)\eta^3\sigma^2}{2n} +\frac{L\tau\eta^2\sigma^2}{2n}.
\label{simplify}
\end{align}\\
For a small $\eta$ that
\begin{align}
1-L\eta -2\tau(\tau-1)L^2\eta^2 \geq 0,
\label{smalleta}
\end{align}
we have
\begin{align}
\mathbb{E}f(\mathbf{w}_{m+1}) &\leq \mathbb{E}f(\mathbf{w}_m)-\frac{\eta}{2}\sum_{t=0}^{\tau -1}\mathbb{E}\Vert \nabla f(\bar{\mathbf{w}}_{m,t})\Vert^2 \nonumber\\
&+\frac{L}{2n^2}\sum_{i \in [n]}\frac{d\cdot BER_m^i\cdot range^2(\Delta \mathbf{w}_m^i)}{3} \nonumber \\
&+\frac{L^2(n+1)\tau(\tau-1)\eta^3\sigma^2}{2n} +\frac{L\tau\eta^2\sigma^2}{2n}.
\label{tosum}
\end{align}\\
Summing (\ref{tosum}) over communication round $m=0,...,K-1$ and multiplying $\frac{2}{K\tau\eta}$ on both sides, we get
\begin{align}
\frac{1}{K\tau}\sum_{m=0}^{K-1}&\sum_{t=0}^{\tau -1}\mathbb{E}\Vert \nabla f(\bar{\mathbf{w}}_{m,t})\Vert^2 \leq  \frac{2(f(\mathbf{w}_0)-f^*)}{K\tau\eta} \nonumber\\
&+\frac{Ld}{3n^2\eta K\tau}\sum_{m=0}^{K-1}\sum_{i \in [n]}BER_m^i\cdot range^2(\Delta \mathbf{w}_m^i)  \nonumber \\
&+\frac{L^2(n+1)(\tau-1)\eta^2\sigma^2}{n}+\frac{L\eta\sigma^2}{n}, 
\label{summed2}
\end{align}\\
which completes the proof of the noisy uplink in Theorem 1.

Then, for FL with downlink bit errors, we just need to modify the notations in Eq.(\ref{add_notation}) to
\begin{align}
&\mathbf{w}_{m+1}=\mathbf{w}_m+\frac{1}{n}\sum_{i\in{[n]}}(\mathbf{w}_{m,\tau}^{i}-\mathbf{w}'_m),\nonumber\\
&\bar{\mathbf{w}}_{m,t}=\frac{1}{n}\sum_{i\in{[n]}}\mathbf{w}_{m,t}^{i},
\label{add_notation1}
\end{align}
and then follow the same derivation process as for the uplink.
\subsection{Proof of Lemmas}
\noindent1) Proof of Lemma 1

Since the effect of BER can be proved to be unbiased, i.e., $\mathbb{E}{(\mathbf{w}')}=\mathbf{w}$, we have
\begin{align}
\mathbb{E}(\mathbf{w}_{m+1})&=\mathbf{w}_m+\frac{1}{n}\sum_{i\in[n]}\mathbb{E}(\mathbf{w}_{m,\tau}^{i}-\mathbf{w}_m)' \nonumber\\
&=\mathbf{w}_m+\frac{1}{n}\sum_{i\in[n]}(\mathbf{w}_{m,\tau}^{i}-\mathbf{w}_m) \nonumber\\
&= \frac{1}{n}\sum_{i\in[n]}\mathbf{w}_{m,\tau}^{i} \nonumber\\
&=\bar{\mathbf{w}}_{m,\tau}.
\label{30}
\end{align}
Because $f$ is $L$-smooth for any variables $\mathbf{x,y}$, we have
\begin{align}
f(\mathbf{w}) \leq f(\mathbf{y}) + \left \langle\nabla f(\mathbf{y}), \mathbf{w}-\mathbf{y}\right \rangle +\frac{L}{2}||\mathbf{w}-\mathbf{y}||^2.
\label{91}
\end{align}
Then, we can have
\begin{align}
f(\mathbf{w}_{m+1}) &\leq f(\mathbf{\bar{w}}_{m,\tau}) + \left \langle\nabla f(\mathbf{\bar{w}}_{m,\tau}), \mathbf{w}_{m+1}-\mathbf{\bar{w}}_{m,\tau}\right \rangle \nonumber \\
&+\frac{L}{2}||\mathbf{w}_{m+1}-\mathbf{\bar{w}}_{m,\tau}||^2.
\label{92again}
\end{align}
Taking the expectation on both sides of (\ref{92again}) and since $\mathbb{E}\mathbf{w}_{m+1}=\bar{\mathbf{w}}_{m,\tau}$ (see (\ref{30})), we have
\begin{align}
\mathbb{E}f(\mathbf{w}_{m+1}) \leq \mathbb{E}f(\mathbf{\bar{w}}_{m,\tau})+\frac{L}{2}\mathbb{E}||\mathbf{w}_{m+1}-\mathbf{\bar{w}}_{m,\tau}||^2.
\label{94}
\end{align}\\
3) Proof of Lemma 3

Using Assumption 1, we have
\begin{align}
\mathbb{E}\Vert&\mathbf{w}_{m+1}-\bar{\mathbf{w}}_{m,\tau} \Vert^2 \nonumber \\
&= \mathbb{E}\Vert \mathbf{w}_m+\frac{1}{n}\sum_{i \in [n]}(\mathbf{w}_{m,\tau}^i-\mathbf{w}_m)'  - \frac{1}{n}\sum_{i \in [n]}\mathbf{w}_{m,\tau}^i \Vert^2 \nonumber\\
&= \mathbb{E}\Vert \frac{1}{n}\sum_{i \in [n]}(\mathbf{w}^i_{m,\tau}-\mathbf{w}_m)'-\frac{1}{n}\sum_{i \in [n]}(\mathbf{w}_{m,\tau}^i - \mathbf{w}_m ) \Vert^2 \nonumber\\
&= \frac{1}{n^2}\mathbb{E}\Vert \sum_{i \in [n]}(\Delta \mathbf{w}^i_m)'-\sum_{i \in [n]}\Delta \mathbf{w}_m^i  \Vert^2 \nonumber\\
&= \frac{1}{n^2}\sum_{i \in [n]}\mathbb{E}\Vert (\Delta \mathbf{w}^i_m)'-\Delta \mathbf{w}_m^i \Vert^2 \nonumber\\
&\leq \frac{1}{n^2}\sum_{i \in [n]}\frac{d\cdot BER_m^i\cdot range^2(\Delta \mathbf{w}_m^i)}{3}.
\label{lemma3p1}
\end{align}

\bibliographystyle{IEEEbib}
\bibliography{strings,refs}

\end{document}